\documentclass{cta-author}

{}
{}
{}

\usepackage{graphicx}
\usepackage[usestackEOL]{stackengine}
\usepackage[noend]{algpseudocode}
\usepackage{mathtools}
\usepackage{algorithm}
\usepackage{graphbox}
\usepackage{url}
\usepackage{xcolor}

\begin{document}


\title{Agile Reactive Navigation for A Non-Holonomic Mobile Robot Using A Pixel Processor Array}

\author{\au{Yanan Liu$^{1,2\corr}$}, \au{Laurie Bose$^{2}$}, \au{Colin Greatwood$^{1}$}, \au{Jianing Chen$^{3}$}, \au{Rui Fan$^{4,5\corr}$}, \au{Thomas Richardson$^{1}$}, \au{Stephen J. Carey$^{3}$}, \au{Piotr Dudek$^{3}$}, \au{Walterio Mayol-Cuevas$^{2,6}$}}

\address{\add{1}{Bristol Robotics Laboratory, The University of Bristol, Bristol, BS16 1QY, United Kingdom}
\add{2}{Visual Information Laboratory, The University of Bristol, Bristol, BS1 5DD, United Kingdom}
\add{3}{Department of Electrical and Electronic Engineering, The University of Manchester, Manchester, M13 9PL, United Kingdom}
\add{4}{The Department of Computer Science and Engineering \& The Department of Ophthalmology, The University of California, San Diego, CA 92093, USA}
\add{5}{Hangzhou ATG Intelligent Technology Co. Ltd., Hangzhou, China}
\add{6}{Amazon, Seattle, USA}
\email{yanan.liu@bristol.ac.uk, rui.fan@ieee.org}}

\begin{abstract}
This paper presents an agile reactive navigation strategy for driving a non-holonomic ground vehicle around a pre-set course of gates in a cluttered environment using a low-cost processor array sensor. This enables machine vision tasks to be performed directly upon the sensor's image plane, rather than using a separate general-purpose computer. We demonstrate a small ground vehicle running through or avoiding multiple gates at high speed using minimal computational resources. To achieve this, target tracking algorithms are developed for the Pixel Processing Array and captured images are then processed directly on the vision sensor acquiring target information for controlling the ground vehicle. The algorithm can run at up to 2000 fps outdoors and 200fps at indoor illumination levels. Conducting image processing at the sensor level avoids the bottleneck of image transfer encountered in conventional sensors. The real-time performance of on-board image processing and robustness is validated through experiments. Experimental results demonstrate that the algorithm's ability to enable a ground vehicle to navigate at an average speed of 2.20 m/s for passing through multiple gates and 3.88 m/s for a `slalom' task in an environment featuring significant visual clutter.
\end{abstract}

\maketitle

\section{Introduction}\label{sec1}

Vision-based mobile robot navigation technology plays an important role in the field of intelligent transportation systems and robotics, being increasingly used across many fields of industry, such as, driverless cars \cite{ozgunalp2017multiple}, assisted living, logistics \cite{ni2016sorting}, and domestic applications \cite{1303538}. However, implementing a mobile robot to perform navigation tasks rapidly, robustly and energy-efficiently can be challenging especially in cluttered environments \cite{liu2018mobile}. A reactive navigation methodology is a solution to this problem that combines sensory information with robot control directly \cite{mujahed2013safe} through processing of sensor data \cite{tobaruela2017reactive}. 

A number of approaches for visual reactive ground navigation have been proposed in recent and past years. In \cite{penin2018vision}, Penin et al. utilised a UAV to perform visual-based reactive navigation, track targets and keep a fixed distance from the target  at a speed of 1.00 m/s. Galluppi et al. \cite{galluppi2014event} developed an autonomous mobile robotic platform using two Dynamic Vision Sensors (DVS) and a SpiNNaker computing system. According to their work, the average navigation speed is less than 0.5 m/s. Many of the existing platforms require expensive computation to perform reactive navigation, which limits reaction time to changing environments and the mobile robot's top speed \cite{lopez2017aggressive,hwu2017self,pan2018agile}. More recent vehicles geared for learning visual navigation, e.g., DeepRacer\footnote{https://aws.amazon.com/deepracer/} or Audi driving cup cars\footnote{https://www.audi-autonomous-driving-cup.com/}, are slow in processing and are yet to demonstrate agile behaviours. Dedicated sensors for car line following do allow faster operation but are tuned for the task, and thus less generic.

\begin{figure}[t!]
\centering
\includegraphics[width=2.3in]{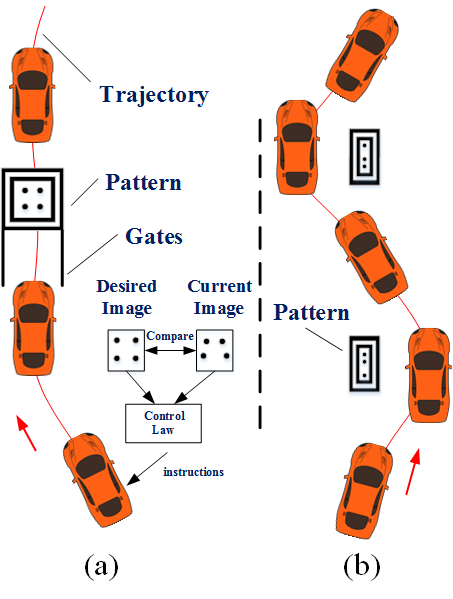}
\caption{Examples of the proposed ground vehicle agile navigation. (a) Passing through the gates. The width of the car body and the gate is 18 cm and 36 cm, respectively; (b) the `slalom' navigation. Experimental video can be seen: \protect\url{https://youtu.be/e85q-yoBuSk} }
\label{fig.overview}
\end{figure}

In agile visual reactive navigation, a vision sensor with a high frame rate is necessary to enable the mobile robot to react to the environment quickly. Various types of visual sensors, such as CCD/CMOS cameras \cite{liu2018design} and RGB-D sensors \cite{7457268} are widely used for map building, robot localisation, navigation and other vision guidance applications \cite{liu2018design,fan2018real,fan2019real,fan2020sne-roadseg}. However, these sensors transmit entire images to a computer for processing, a process that requires relatively significant time and power, decreasing the response time of the robot \cite{greatwood2018perspective}. The key contribution of this paper compared to our previous work \cite{Greatwood2017} includes improving iterative flooding operations to enhance its robustness when extracting disks; designing a filtering algorithm based on the neighbourhood pixel values to denoise the image; exploiting the PID control to complete the navigation task and integrating the improvements mentioned above to directly link the visual information from the image plane to robot control instructions enabling agile reactive navigation. Compared to a traditional visual sensor, each pixel in the Pixel Processor has storage and processing abilities, which supports fast and lower-power parallel computation directly on the visual sensor. The Pixel Processor Array (PPA) is capable of performing general-purpose vision algorithms at several thousand frames per second due to its parallel computing features \cite{Carey2013}, \cite{Bose2017}. This paper presents a novel agile reactive navigation strategy using the PPA and a mobile robot with a low-priced single-board computer and a simple control structure. In our applications, a car-like mobile robot self-navigates, running through multiple gates at an average speed of 2.20 m/s according to the guidance provided by visual information extracted from the pre-designed patterns pasted on the gates (see Fig. \ref{fig.overview}(a)). The control instructions for the mobile robot are generated by comparing the desired image and the current image. The mobile robot also achieves a highest speed of 3.88 m/s during a `slalom' process (Fig. \ref{fig.overview}(b)). In particular, the mobile robot is directly instructed by the coordinates of feature points on the patterns themselves, hence there is no need to perform  transformation calculations among camera coordinate system, image coordinate system and ground vehicle coordinate system. 

The remaining sections of this paper are organised as follows: Section 2 describes the proposed vision system including the introduction of the SCAMP-5 vision system and image processing algorithms. In Section 3, we present the robot control system which consists of hardware and software structure. Section 4 illustrates the agile reactive navigation method. Experimental results are illustrated and the proposed system performance is evaluated in Section 5. Finally, Section 6 summarises the paper and provides recommendations for future work.

\section{Vision System Description}

This section introduces the SCAMP-5 vision system, compares it with the traditional visual sensor and illustrates the gate pattern recognition algorithms based on the SCAMP-5.

\subsection{SCAMP-5 Vision System}

\begin{figure}[t!]
\centering
\includegraphics[width=3.2in]{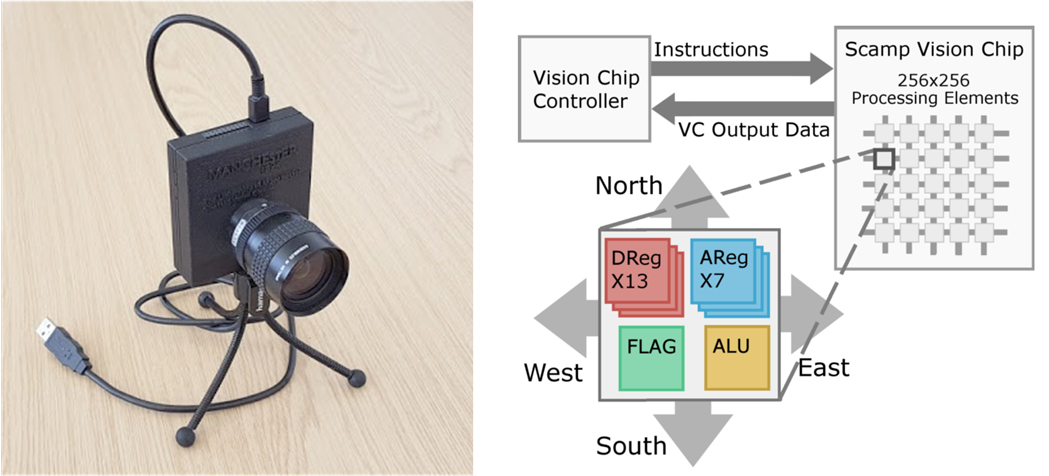}
\caption{SCAMP-5d (Left) and the processing elements in Pixel Processor Arrays (Right) \cite{Bose2017}.}
\label{fig_PE}
\end{figure}

SCAMP-5d (see Fig. \ref{fig_PE}) is the latest version of the programmable, general-purpose, power-efficient and high-performance pixel processor vision chip family developed by the University of Manchester \cite{Carey2013}.  
The PPA of the SCAMP-5 vision chip  integrates a massively parallel Single Instruction Multiple Data (SIMD) processor array into the image sensor pixels. It is the combination of the image sensor's pixels and processor granting it the ability of both a traditional visual sensor and a programmable processor. The vision sensor of the SCAMP-5 system consists of a grid of 65,536 (256 $\times$ 256) processing elements (PE). Each PE in the image array consists of 7 analogue registers, 13 digital registers, and 1 digital `FLAG' register, which enables this vision system to store 7 grey-value images, 13 binary images and to support conditional execution among frames. Moreover, arithmetic operations, logic operations and neighbourhood operations on the image plane are available.

\begin{figure}[t!]
\centering
\includegraphics[width=3.2in]{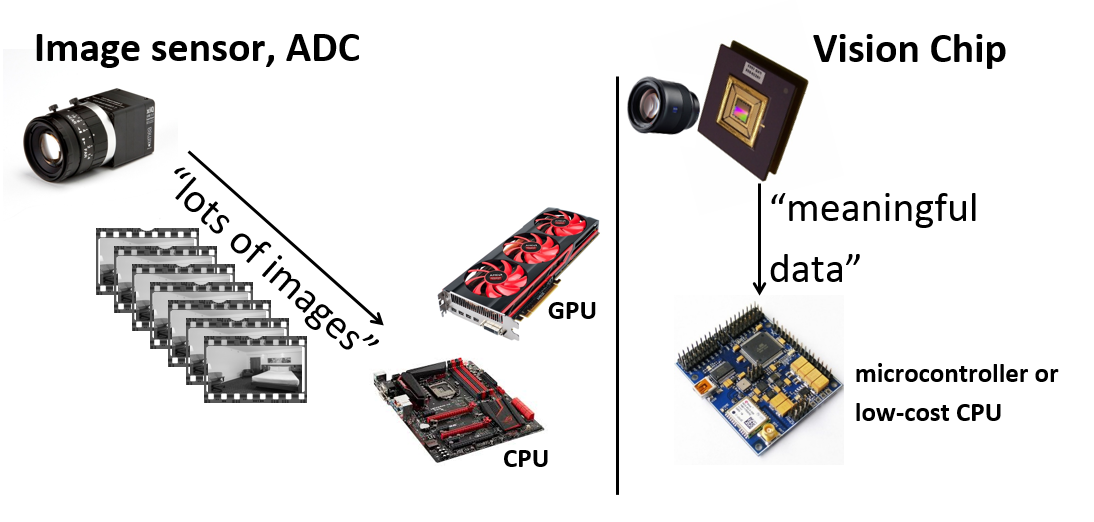}
\caption{Comparison of the workflow between the traditional vision system and the SCAMP-5 vision system.}
\label{fig_scamp5_workflow}
\end{figure}

Unlike the conventional image sensor which only contains the light sensitive pixel circuits, SCAMP-5 features the integration of image  sensors and processors on a single silicon die. By doing so, the image processing happens directly on the image plane. Thus, the image processing results rather than the image itself can be transferred to other controllers (see Fig. \ref{fig_scamp5_workflow}). For most tasks, the data/calculations resulting from this image processing are far smaller than the raw image data, which greatly reduces the transfer time between SCAMP-5 and the controller compared to a standard camera. The vision chip controller is based on the ARM-Cortex-M, it supports programming and general-purpose computation tasks. The image processing algorithms can be developed using a standard C++ compiler.
on the vision chip controller. After compiling the program, the image processing kernels for acceleration on the processor array are sent to each PE to perform processing tasks. These PEs are able to perform basic computation tasks in parallel, hence enabling fast-speed on-sensor image processing. Finally, the image processing results are sent back to the chip controller and then transferred to next-level hardware, such as the single board computer through GPIO or USB interface. Moreover, the SCAMP-5 system enables real-time image processing on the focal plane while consuming minimal power (between 0.1W and 2W depending on the algorithm), which is suitable for embedded systems in the field of robotics \cite{Bose2017,Greatwood2017,martel2016real,Chen2017,Chen2020icra}.

\subsection{Gate Pattern Recognition Algorithm}

This subsection describes the SCAMP-5 algorithm used to detect the pre-designed patterns. Patterns are a common means to give instructions to human drivers or even to autonomous cars such as in line following. The challenges include tolerance to noise, clutter and fast enough processing. The patterns we use are primarily aimed at demonstrating agile visuo-control behaviours.

\begin{figure}[t!]
\centering
\includegraphics[width=3.4in]{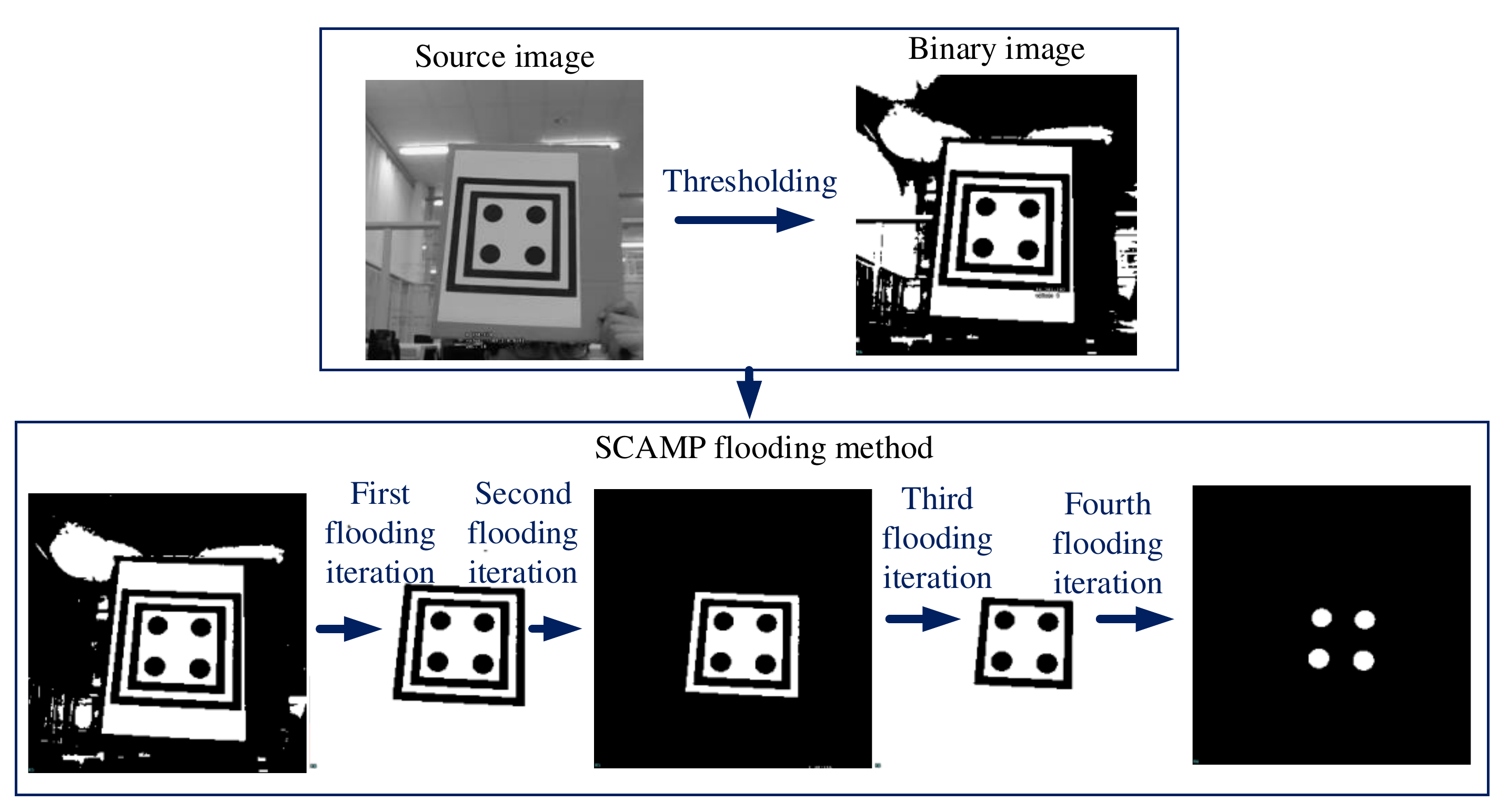}
\caption{Pattern extraction procedure. The four extracted dots in the final result form a quadrangle.}
\label{fig_imageprocessing}
\end{figure}

\begin{figure}[t!]
\centering
\includegraphics[width=3.3in]{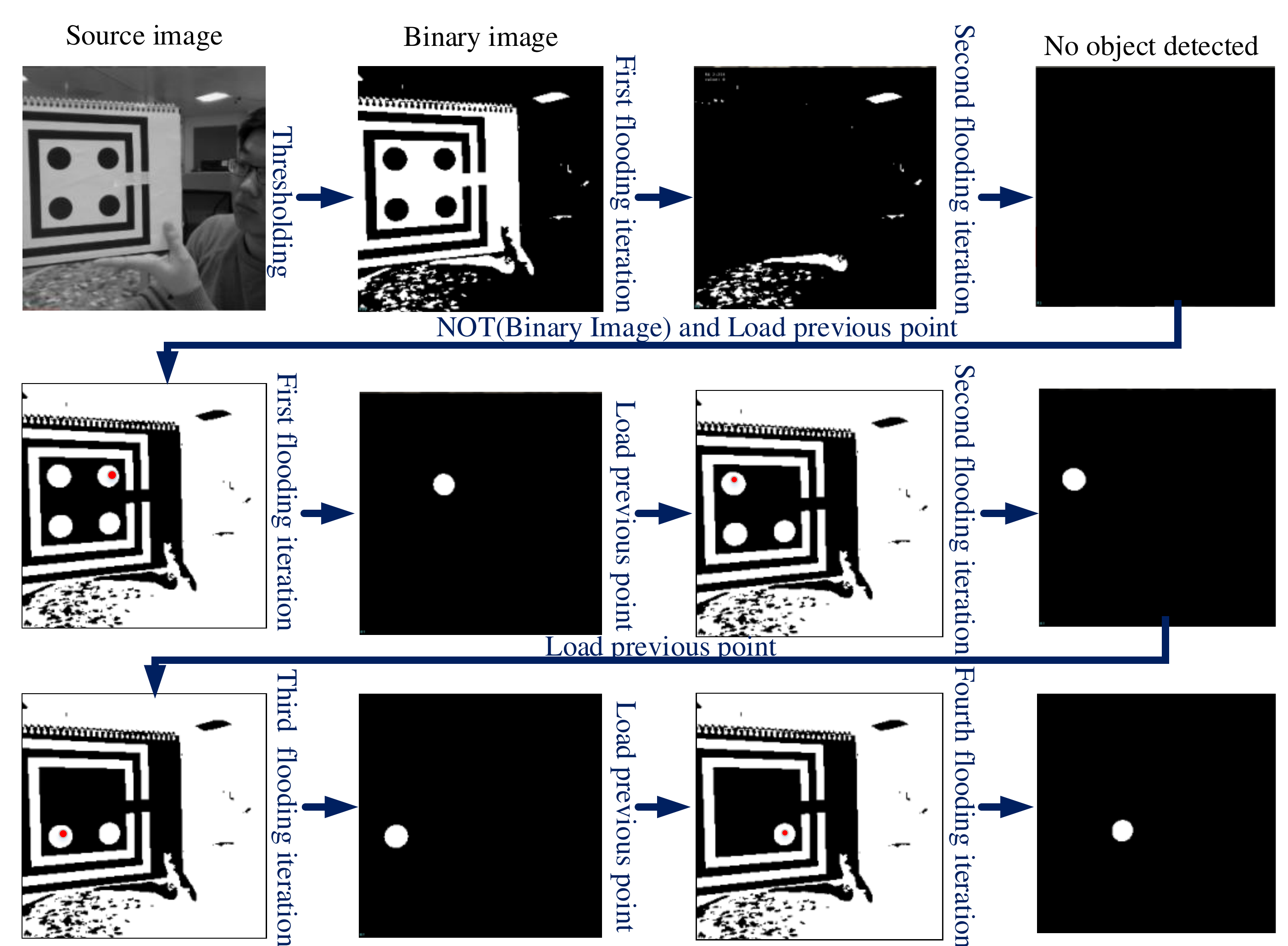}
\caption{Image processing procedure in which the squares are broken.}
\label{fig_imageprocessing2}
\end{figure}

For the gate's task, each pattern contains four black disks surrounded by two black concentric squares. The disk coordinates in the image plane are utilised to adjust the rover's position and orientation, guiding it to go through multiple gates at high speed. Fig. \ref{fig_imageprocessing} and Fig. \ref{fig_imageprocessing2} show the image processing procedure conducted on SCAMP-5. High-frame-rate real-time image processing is typically challenging to conduct with a standard camera and computer setup due to the limited rate of image capture and delay in data transfer that such devices have. In this work the target extraction algorithm is designed to exploit SCAMP-5's fast image flooding ability to minimise computation time per image and improve the image processing robustness. Iterative flooding operations are used to extract centres of four disks out of the cluttered background \cite{Greatwood2017}. This method fills regions with value `1's except these areas closed by value `0's. Hence those areas that are out of the closed regions would be filled with `0's and with this method, the background of the pattern is easily eliminated within several iterations. As shown in Fig. \ref{fig_imageprocessing}, after the first flooding iteration, the background is eliminated. Therefore, the time cost for this kind of operation is much less than that of the traditional camera, because this flooding operation is carried out in all PEs in parallel and asynchronously rather than pixel by pixel for traditional image processing.

The extraction of four points in the pattern relies on the presence of two black concentric boundaries. However, issues, such as target occlusion, bright light reflections or simply the target being distant from the camera, may break or merge together those two boundaries, causing failure of the dot extraction process. For example, in Fig. \ref{fig_imageprocessing2}, the pattern in the source image is partly outside the view field and the boundaries broken on the right hand side for demonstration. In this case, the direct flooding method is not useful since four disks are not enclosed by `0's. To improve the robustness of the pattern extraction, the prior knowledge of where the dots were located in the previous frame is used whenever the two black boundaries are not present. This is illustrated in Fig. \ref{fig_imageprocessing2}, when there is no object remaining after two flooding and inversion operations. Firstly, the inversion of the current binary image is performed. Then, a point from the last detected disk centre is loaded into the current frame. Since the frame rate of the SCAMP-5 is set to 200 fps or more,
the shift between two consecutive frames is small, and thus, we assume that the loaded point falls into the current corresponding disk. Then, flooding is conducted outwards from this point using the current inverted image as a mask. If the point from which flooding was performed existed within the disk, the resulting image contains only that disk which can then be easily extracted. After getting rid of the first extracted disk from the inverted image, the location of the remainder of the disks can be obtained using the loading point and flooding method iteratively. The detailed algorithm description can be seen in Algorithm \ref{alg:image processing}.

\begin{figure}[t!]
\centering
\includegraphics[width=3.3in]{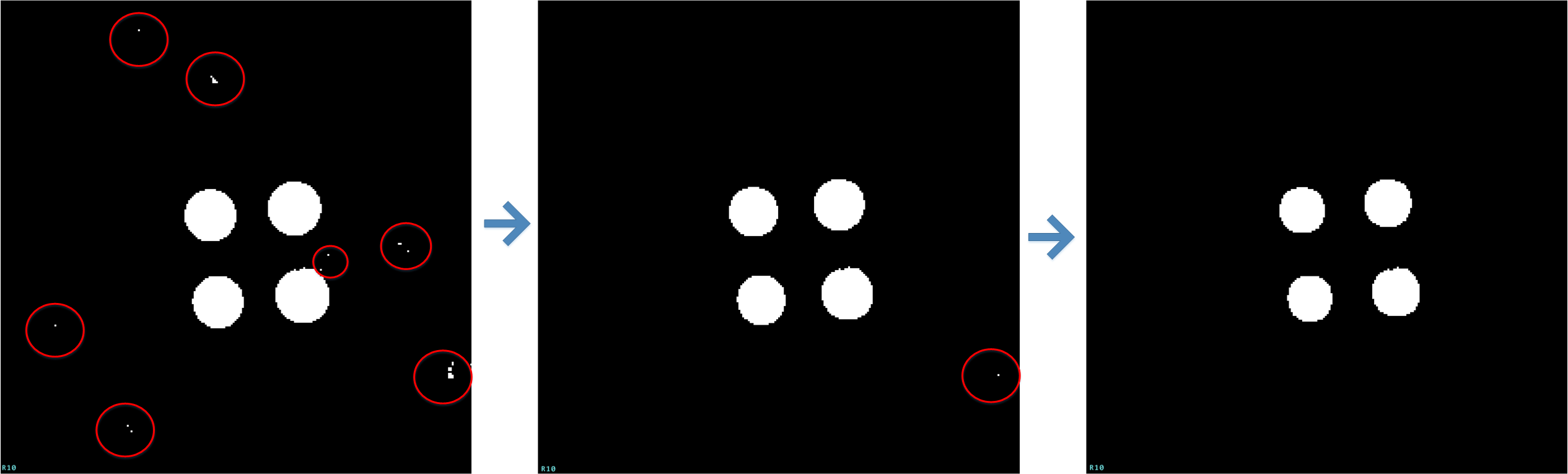}
\caption{Image de-noising, the shrink radius is 2 pixels. Noisy pixels are circled in red. All noisy pixels are eliminated in the third image.}
\label{fig_denoise}
\end{figure}

\begin{algorithm}
\caption{Disk extraction based on the SCAMP-5}\label{alg:image processing}
\begin{algorithmic}
\State \textbf{INPUT}: 
  \State $Threshold$       // used to get binary image
  \State $Iterations$ // iteration times for flooding and inversion operation
  \State $Disc\underline{~}num$    // pre-set number of disks
\State  \textbf{OUTPUT}:
  \State $Coordinates$ // coordinates of four disks' centres
\While {true}
 \State \textit{R5 = Scamp5\underline{~}get\underline{~}image()}
 \State \textit{R6 = Scamp5\underline{~}threshold(R5,Threshold)}
 \For{ \textit{n = 0 to Iterations}}
    \State \textit{R7 = Scamp5\underline{~}flood(R6)}
    \State \textit{R6 = AND(R6, NOT(R7))}
    \State \textit{R6 = NOT(R6)}
    \EndFor
 \State \textit{Image\underline{~}filtering (see Algorithm \ref{alg:image filtering})}
 \State \textit{Num = Object\underline{~}detect(R6)}
  \If{\textit{Num == Disc\underline{~}num}}
     \State \textit{Coordinates = Scamp5\underline{~}scan\underline{~}boundingbox(R6)}
 \Else
     \State \textit{R7 = Scamp5\underline{~}load\underline{~}point(previous\underline{~}point)}
     \State \textit{R7 = Scamp5\underline{~}flood(R6)}
     \State \textit{R8 = R6 XOR R7}
     \State \textit{R6 = R8}
     \State \textit{Coordinates = Scamp5\underline{~}scan\underline{~}boundingbox(R7)}
 \EndIf
 \State \textit{previous\underline{~}point = Coordinates}
\EndWhile
\State \algorithmicreturn{ \textit{Coordinates}} 
\end{algorithmic}
\end{algorithm}

Noise in an image is caused by various factors, such as overexposure, cluttered environment and quantisation. In this scenario, the light spots caused by overexposure or reflection are the main sources of noise. To eliminate the noise, an image filtering method based on the neighbour communication is conducted upon SCAMP-5. Since each processing element can communicate with its four neighbours (north, south, west, and east), a given pixel can access its neighbours' data directly. Four new pictures are obtained after moving the source image into four directions with a pixel distance $P\textsubscript{step}$. Then using `AND' operator to add these four images together on the image plane directly. In Fig. \ref{fig_denoise}, noisy area whose radius is smaller than shrink radius will be eliminated. The pseudo codes are given in Algorithm \ref{alg:image filtering}.

\begin{algorithm}
\caption{Image filtering based on the SCAMP-5}\label{alg:image filtering}
\begin{algorithmic}
\State \textbf{INPUT}: \textit{P\textsubscript{step}}
\State \textbf{OUTPUT}: $R10$
\While {true}
    \State \textit{R5 = Scamp5\underline{~}get\underline{~}image()}
    \State \textit{R6 = Move R5 to the north with $P\textsubscript{step}$}
    \State \textit{R7 = Move \textit{R5} to the south with $P\textsubscript{step}$}
    \State \textit{R8 = Move \textit{R5} to the west with $P\textsubscript{step}$}
    \State \textit{R9 = Move \textit{R5} to the east with $P\textsubscript{step}$}
    \State \textit{R10 = $\textit{R6} \cap \textit{R7} \cap \textit{R8} \cap \textit{R9}$}
\EndWhile
\State \algorithmicreturn{ \textit{R10}} 
\end{algorithmic}
\end{algorithm}

\begin{figure}[!t]
\centering
\includegraphics[width=3.3in]{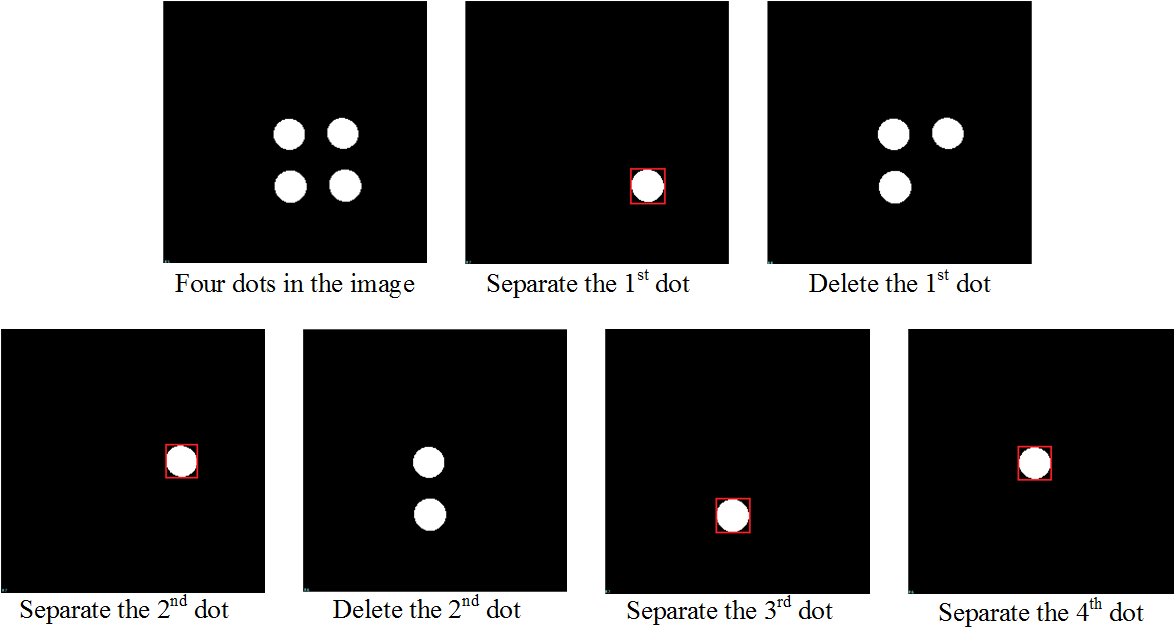}
\caption{An example of getting coordinate of each dot's centre.}
\label{fig_point_coordinate}
\end{figure}

We now discuss the way of getting each dot's centroid in the image plane. First an inbuilt function $scan\underline{~}event$ of the SCAMP-5 is used to find the location of a white pixel in the image. This white pixel must be located within one of the disks we wish to extract. We then perform a flooding operation originating from this extracted point, using current image as a mask, resulting in an image consisting of the entire disk the point was within, as shown in the second image of Fig. \ref{fig_point_coordinate}. The bounding box of all white pixels in the image (which in this case are those of the disk) is then extracted using another built-in SCAMP-5 function $scan\underline{~}boundingbox$. After that, the disk is removed from the original binary image using a NOT operation. With this method, one white disk can be extracted. This process is similar to that used in Fig. \ref{fig_imageprocessing2} where squares are not intact. The $scan\underline{~}boundingbox$ function outputs the centre position of the white region. After performing this type of operation for four iterations, all centre positions of the disks can be extracted separately. 

The image processing time of all algorithms mentioned above is recorded by the SCAMP-5 application that receives the image processing information from the SCAMP-5 vision system through a USB cable. In this scenario, the frame rate is set to 200 fps given the illumination conditions in the indoor arena. It is noteworthy that the proposed image processing algorithm can be easily performed at more than 2000 fps with enough light illumination.

\section{Architecture of Robot System}

\begin{figure}[!h]
\centering
\includegraphics[width=3in]{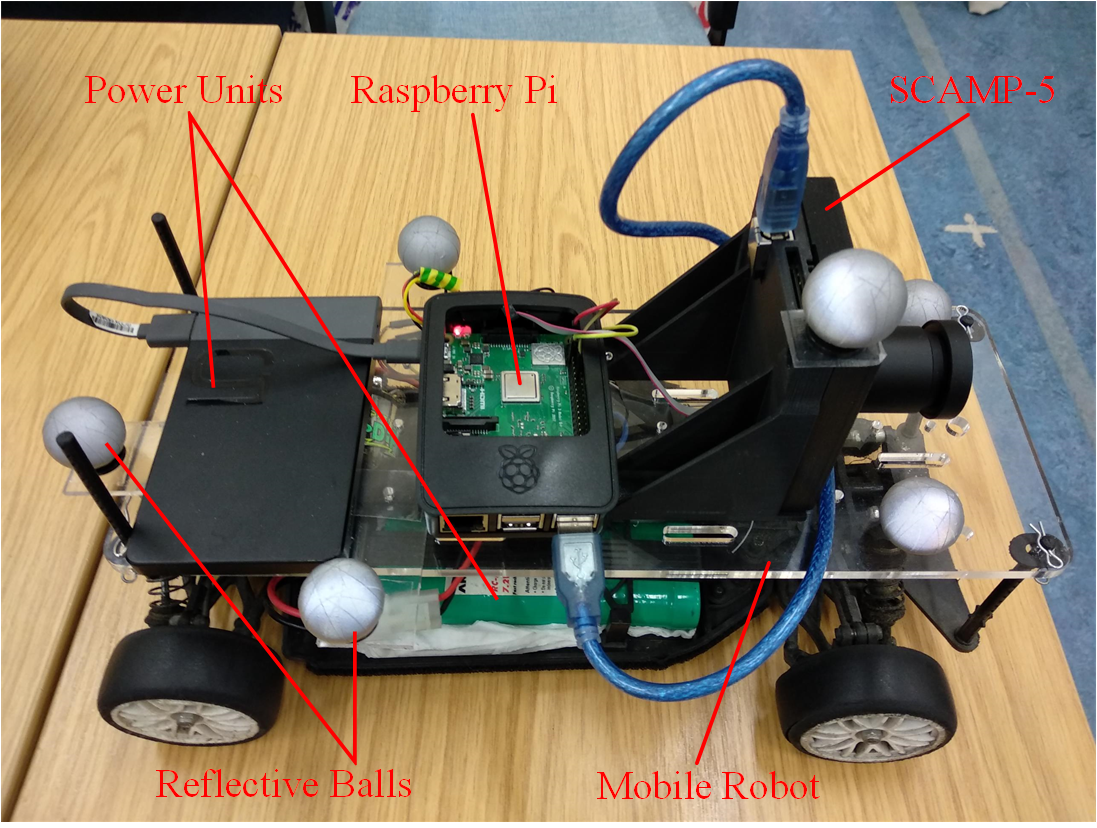}
\caption{Subaru RC car (1:10 scale model) and hardware used in experiments.}
\label{fig_system_hardware}
\end{figure}

\begin{figure}[!h]
\centering
\includegraphics[width=3in]{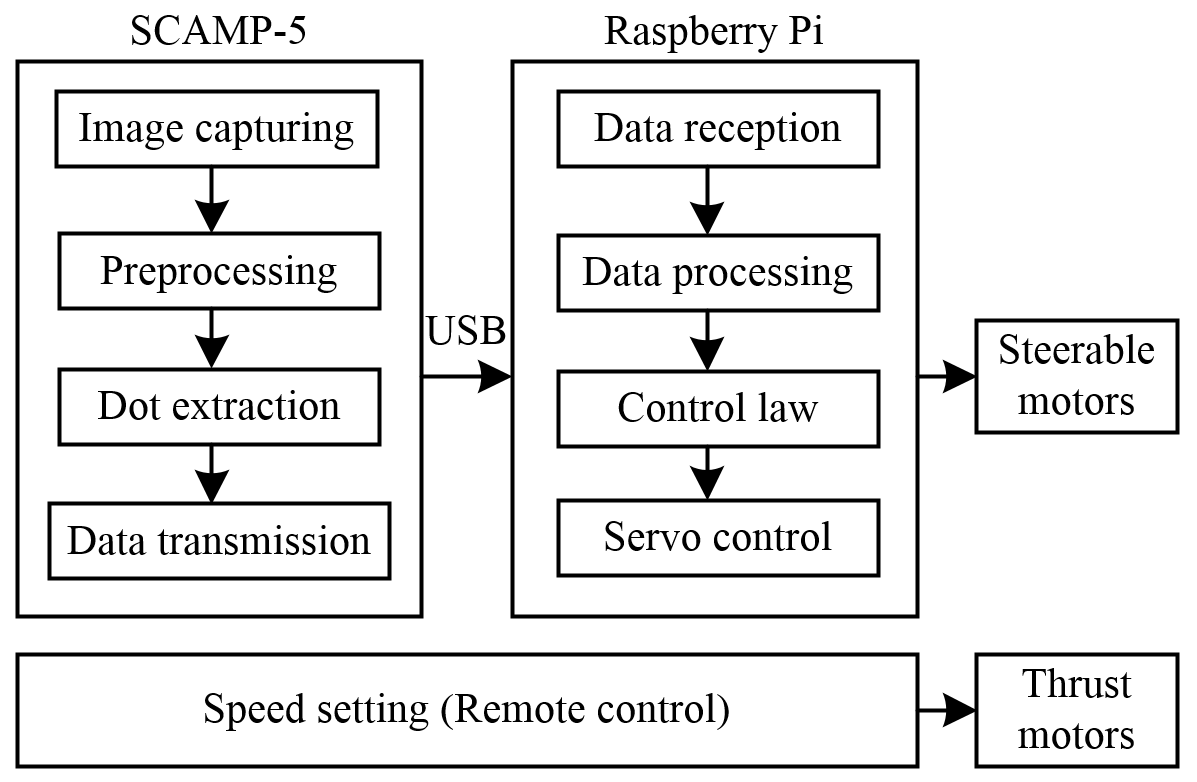}
\caption{Control architecture of the navigation system.}
\label{fig_software_architecture}
\end{figure}

A Subaru rover chassis, shown in Fig. \ref{fig_system_hardware}, was utilised for the agile navigation experiment. The rover was controlled by both the visual data from the SCAMP-5 and the remote control shown in Fig. \ref{fig_software_architecture}. Specifically, the visual data controls the steering while the speed of the car is set to a fixed value using the remote control. As shown in Fig. \ref{fig_software_architecture}, images are captured and processed on the vision chip and the processing results are sent back to the chip controller. Finally, these coordinates are transferred to the Raspberry Pi, which handles read-out data and performs mobile robot navigation tasks. The detailed hardware and its hardware configurations can be seen in Table \ref{table_hardware}.

\begin{table}[!t]
\processtable{LIST OF HARDWARE ADOPTED FOR THE NAVIGATION SYSTEM.\label{table_hardware}}
{\begin{tabular*}{20pc}{@{\extracolsep{\fill}}lll@{}}\toprule
Hardware name    & Hardware configuration \\
\midrule
Subaru RC Car & SKU H94123-12344 \\
SCAMP-5 Vision System  &SCAMP-5d (gray-level image, 256$\times$256 array)\\
Raspberry Pi  & Version 3 Model B+, CPU: 1.4GHz \\
Subaru power unit & ANSMANN, RC-Racing Pack \\
Raspberry Pi power unit & HUAWEI Colorphon 5 \\
Remote control & TARANIS \\
\botrule
\end{tabular*}}{}
\end{table}

\section{Agile Reactive Navigation}

\subsection{Object Tracking}

The SCAMP-5 extracts the four disks located within any visible targets which are then utilised to steer the mobile robot through the gate with the observed target. Controls are generated by comparing the four disks extracted from a pattern in view against four reference disk for a pattern at a known distance and angle. However, when carrying out the disk extraction, the order of these four dots is unknown because the relative position between the camera and the gate can be random. Hence, a method is developed to make these eight points correspond to each other before making a comparison.

\begin{figure}[!t]
\centering
\includegraphics[width=2.8in]{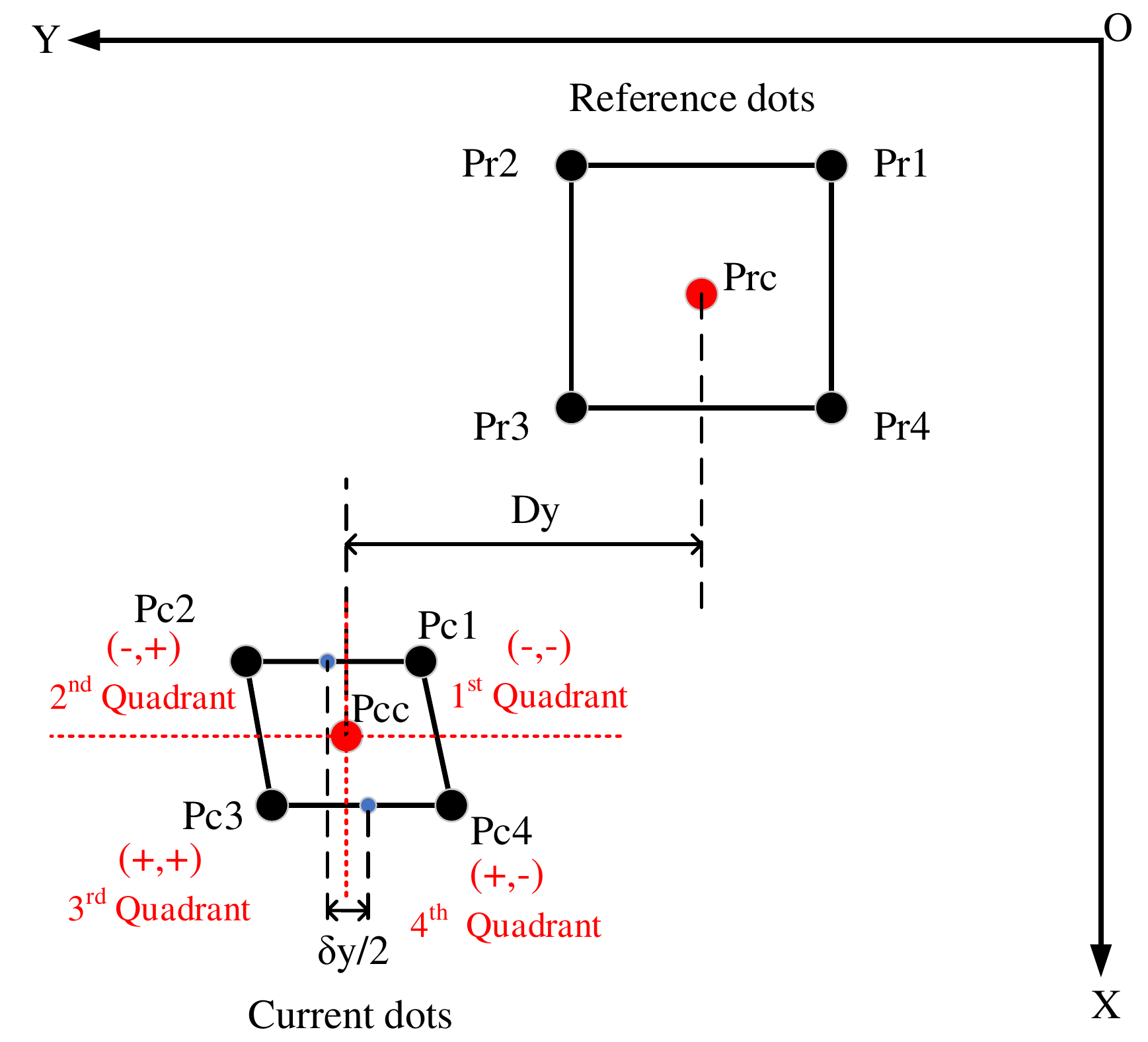}
\caption{Quadrant allocation for four current dots and its difference from the reference dots. These dots represent the centres of extracted disks and they form a rectangle in reference image and a quadrangle in current image respectively.}
\label{fig_reference_points}
\end{figure}

As shown in Fig. \ref{fig_reference_points}, the centre of these four dots $P\textsubscript{cc}$ can be obtained by $P\textsubscript{cc} = \sum P\textsubscript{ci}/4$. By comparing the difference between $P\textsubscript{cc}$ and $P\textsubscript{ci}$, these four dots can be allocated into the corresponding quadrant. The current picture is divided into four quadrants each of which will contain a specific disk which can then be compared to the corresponding disk in a reference pattern.

\begin{equation}\label{eq_sign}
	Sign = P\textsubscript{ciy} - P\textsubscript{ccy}, \quad i \in \{1,2,3,4\}.
\end{equation}

Where $Sign$ represents the sign symbol of the difference between $P\textsubscript{ccy}$ and $P\textsubscript{ciy}$ along $y$ axis.

\subsection{Mobile Robot Reactive Navigation}

This subsection compares the difference in both position and shape between the reference rectangle and the currently processed quadrangle to generate instructions for the mobile robot to run through gates vertically. The reference image is defined with an image that was pre-set by putting the ground vehicle in front of the gate at a distance which is the minimum distance that is in sight of the sensor. The reference rectangle is compared to the currently processed quadrangle; from this comparison, controls are generated for the RC rover which will bring the observed rectangle closer to the reference, guiding it towards the gate. As the pose of the rover gets closer to that at which the reference image was taken, the similarity between the quadrangles correspondingly increases. This control guides the rover through the gate without collision enabling the rover to see the next gate.

During reactive navigation, the distance $D\textsubscript{y}$ between $P\textsubscript{cc}$ and $P\textsubscript{rc}$, the deformation $\delta y$ along y axis (see Fig. \ref{fig_reference_points}) act as the input for the PID control. 

\begin{equation}\label{eqn_d}
	D\textsubscript{y} = P\textsubscript{cc} - P\textsubscript{rc},
\end{equation}
where $D\textsubscript{y}$ is utilised to adjust the angle of the front wheels towards the target. $\delta y$ describes the deformation of the pattern along y axis caused by the relative position between the pattern and the rover. In this scenario, $\delta y$ is the difference between top side centre and bottom side centre in the quadrangle. 

\begin{equation}\label{eqn_deltay}
	\delta y = P\textsubscript{c3y} + P\textsubscript{c4y} - (P\textsubscript{c1y} + P\textsubscript{c2y}),
\end{equation}
 where $\delta y$ is used as a compensation to slightly change the wheel angle and attempt to take the rover straight through the gate rather than going through at an angle. The adopted control method is as follows:

\begin{equation}\label{eqn_control}
    Output =  Output_1 + Output_2 ,
\end{equation}

\begin{equation}\label{eqn_control1}
    Output_i = K_{Pi}e(k)_i + K_{Ii}\sum_{j=0}^{k}e(j)_i + K_{Di}[e(k)_i - e(k-1)_i]
\end{equation}

\begin{equation}\label{eqn_control2}
    e(k)_1 = D\textsubscript{y} ,
\end{equation}

\begin{equation}\label{eqn_control3}
  e(k)_2 = \delta y .
\end{equation}

Where, $i \in \{1,2\}$, $K_P$, $K_I$ and $K_D$ are coefficients adjusted experimentally in this paper using Ziegler-Nichols \cite{ziegler1942optimum}. $e(k)_i$ is the error between $D\textsubscript{y}$, $\delta y$ and 0. As we can see from the equation, the visual information is slightly processed before generating control instructions for the mobile robot. The aim of this PID control is to minimise both $D\textsubscript{y}$ and $\delta y$ to 0. During this process, the ground vehicle is moving towards the gate while adjusting its pose to enter the gate head on.

\section{Experimental Results}

\begin{figure}[t!]
\centering
\includegraphics[width=3in]{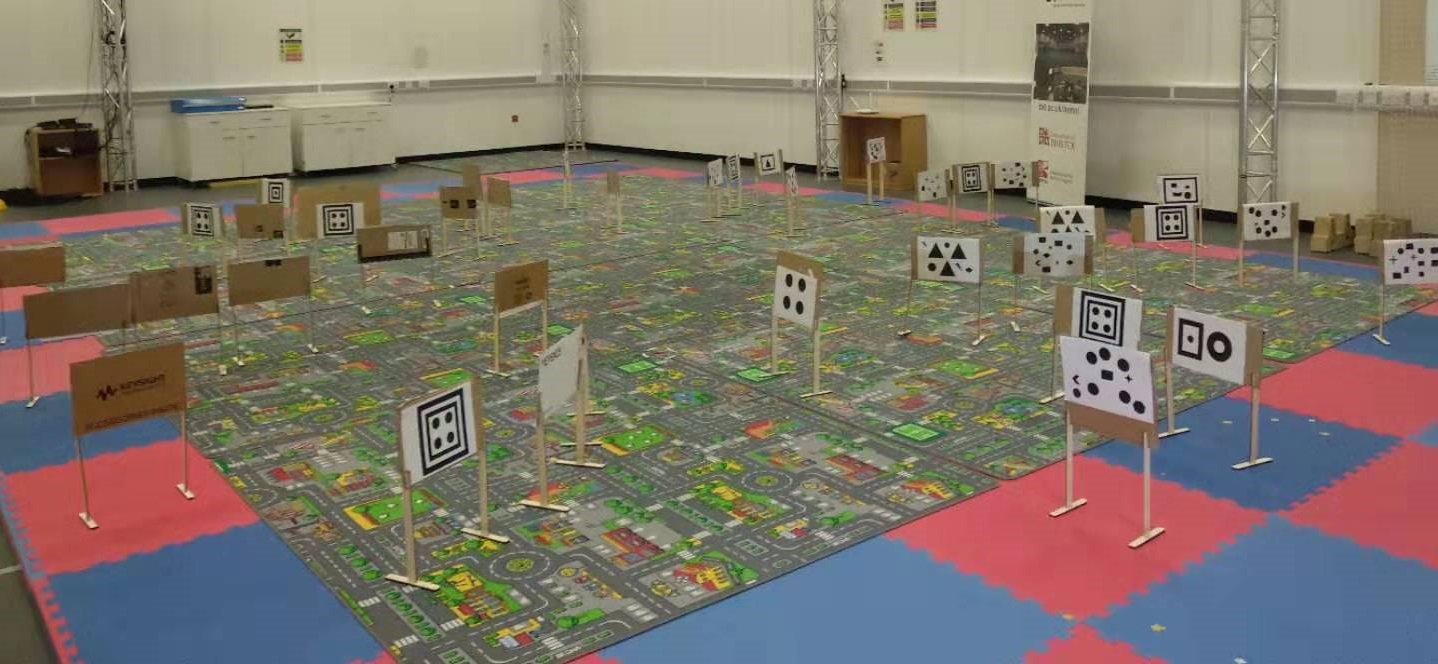}
\caption{Gates and clutter layout in the arena.}
\label{fig_gate_layout}
\end{figure}

Experiments are carried out to test how agile the ground vehicle would track the target (the gate) and run through the gate in heavily cluttered environment. The mobile robot is set at a constant speed using the remote control. In Fig. \ref{fig_gate_layout}, there are eight gates placed in the robot arena taking into consideration space limitations and the vehicle's turning radius. Beside each gate, there are some other gates with a random or similar pattern to the real pattern acting as a disturbance to show the robustness of this navigation system to the cluttered environment. 

\begin{figure}[!t]
\centering
\includegraphics[width=3.2in]{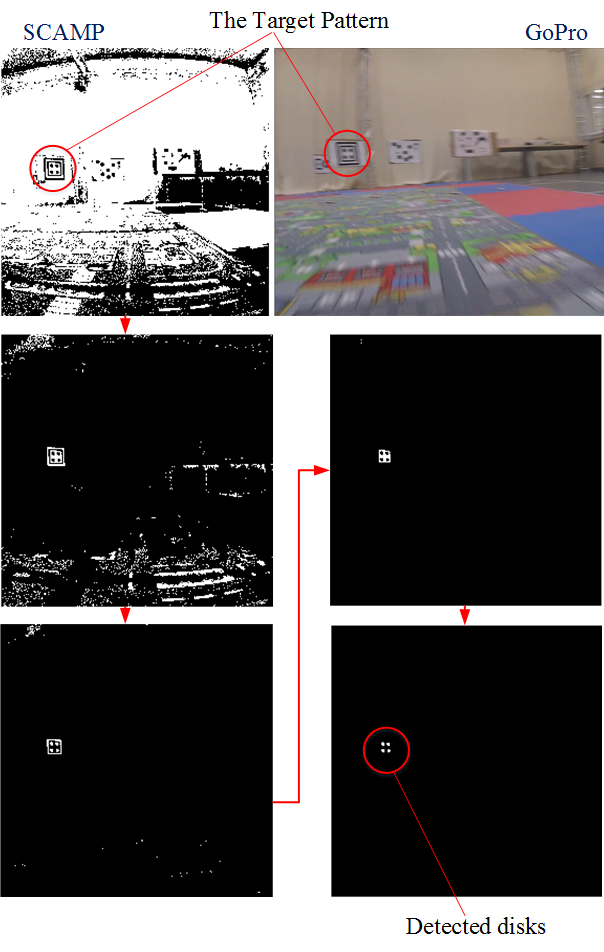}
\caption{Image captured by the SCAMP-5 and GoPro during high-speed navigation. The SCAMP-5 detects four disks during high-speed navigation in a cluttered environment.}
\label{fig_image_comparison}
\end{figure}

\begin{figure}[!h]
\centering
\includegraphics[width=3.5in]{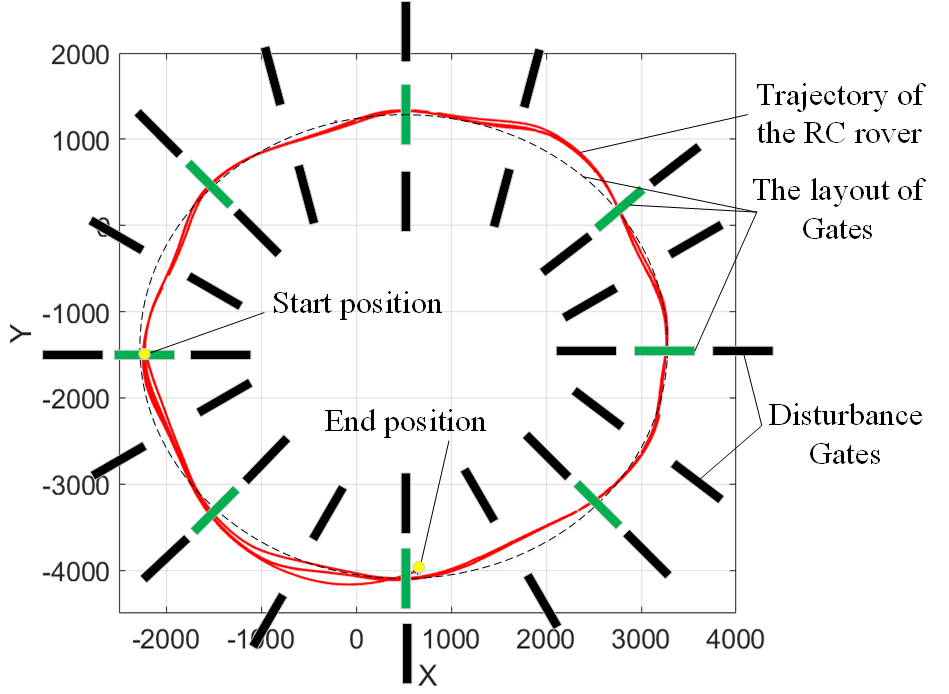}
\caption{Trajectory in the arena.}
\label{fig_trajectory_image_plane}
\end{figure}

\begin{figure}[t!]
\centering
\includegraphics[width=2.2in]{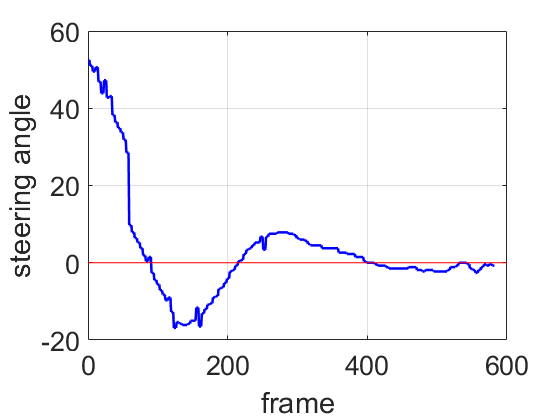}
\caption{The angle change when the rover is approaching a gate.}
\label{fig_pid_control_position}
\end{figure}

\begin{figure}[t!]
\centering
\includegraphics[width=3in]{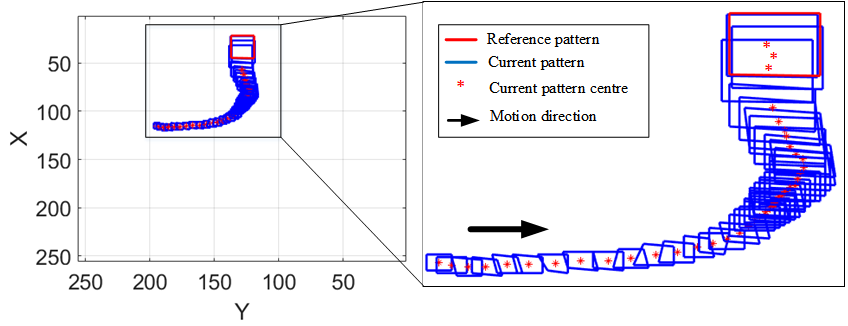}
\caption{Trajectory in the image plane.}
\label{fig_trajectory_image_plane_2}
\end{figure}

 The views in front of the rover are recorded by a GoPro camera. As we can see from Fig. \ref{fig_image_comparison}, the images captured by the GoPro are blurred because of the high speed and vibration of the chassis during navigation. However, the SCAMP is able to get clear images and output effective visual data. When the mobile robot is running through these gates, its trajectory is recorded by the VICON capture system. Fig. \ref{fig_trajectory_image_plane} shows the rover's paths and the layout of eight gates and the disturbance. As we can see, the rover runs though these gates, which indicates the effectiveness of the proposed control method and the robustness of the image processing. Fig. \ref{fig_pid_control_position} shows the changes of the front wheel angle when the rover is approaching a gate and the steering angle is asymptotically close to 0. The extracted dots are recorded during the navigation and plotted. As shown in Fig. \ref{fig_trajectory_image_plane_2}, the relative position of these four dots is changing when the rover is running towards a gate. The difference between the reference image and the current image is used for controlling the motion of the mobile robot. Consequently, in the image plane, their difference should be getting increasingly smaller during the navigation process. The expected phenomenon can be seen in Fig. \ref{fig_trajectory_image_plane_2}, where the current pattern shape is becoming increasingly closer to the reference pattern, in terms of both position and shape.
 
Although all the patterns are designed identically, their positions on the gate are slightly different. In addition, the view of the SCAMP-5 is possibly not horizontal because of its suspension system during high-speed motion. As a result, the final image when the mobile robot is about to pass through the gate could be slightly different from the reference image. 

\begin{figure}[t!]
\centering
\includegraphics[width=2.2in]{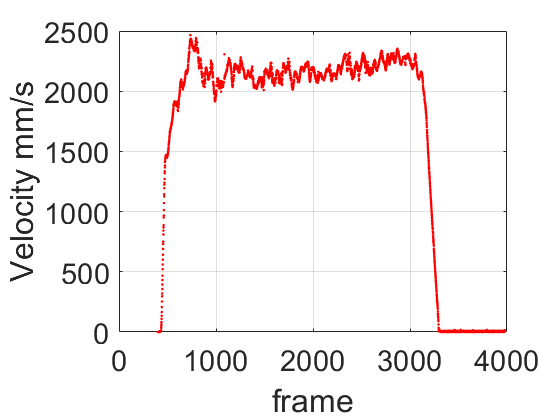}
\caption{Velocity curve for the whole navigation process recorded by VICON.}
\label{fig_trajectory_8_gates}
\end{figure}

The agility of this reactive navigation can be seen from Fig. \ref{fig_trajectory_8_gates}. The average navigation speed is calculated 2.20 m/s and the maximum speed reaches 2.50 m/s according to the VICON tracking data, which shows the rover response to the environment quickly with the guidance of the SCAMP-5.

The time cost for each procedure of the whole control system, including that in SCAMP-5 and Raspberry Pi, is measured and listed in Table \ref{table_time_cost}. Our SCAMP-5 pipeline is able to perform at over 2000 fps according to Table \ref{table_time_cost}. This accounts for the image and PID processing but does not include the image exposure time which will depend on environmental illumination conditions.

\begin{table}[!t]
\processtable{\textsc{Time cost for different components of the system.}\label{table_time_cost}}
{\begin{tabular*}{20pc}{@{\extracolsep{\fill}}lll@{}}\toprule
Processing Steps    & Time Cost ($\mu s$) \\
\midrule
Image capturing and thresholding & 52 \\
Flooding method to extract dots & 63  \\
(including normal and broken squares) & \\
Image de-noising  & 4 \\
Getting coordinates of four dots' centre & 328 \\
Coordinates transmitted to Raspberry Pi & $\approx$ 24 \\
PID control in Raspberry Pi & $\approx$ 1   \\
\midrule
Total &  $\approx$ 472 \\
\botrule
\end{tabular*}}{}
\end{table}

\begin{figure}[t!]
\centering
\includegraphics[width=1.0in]{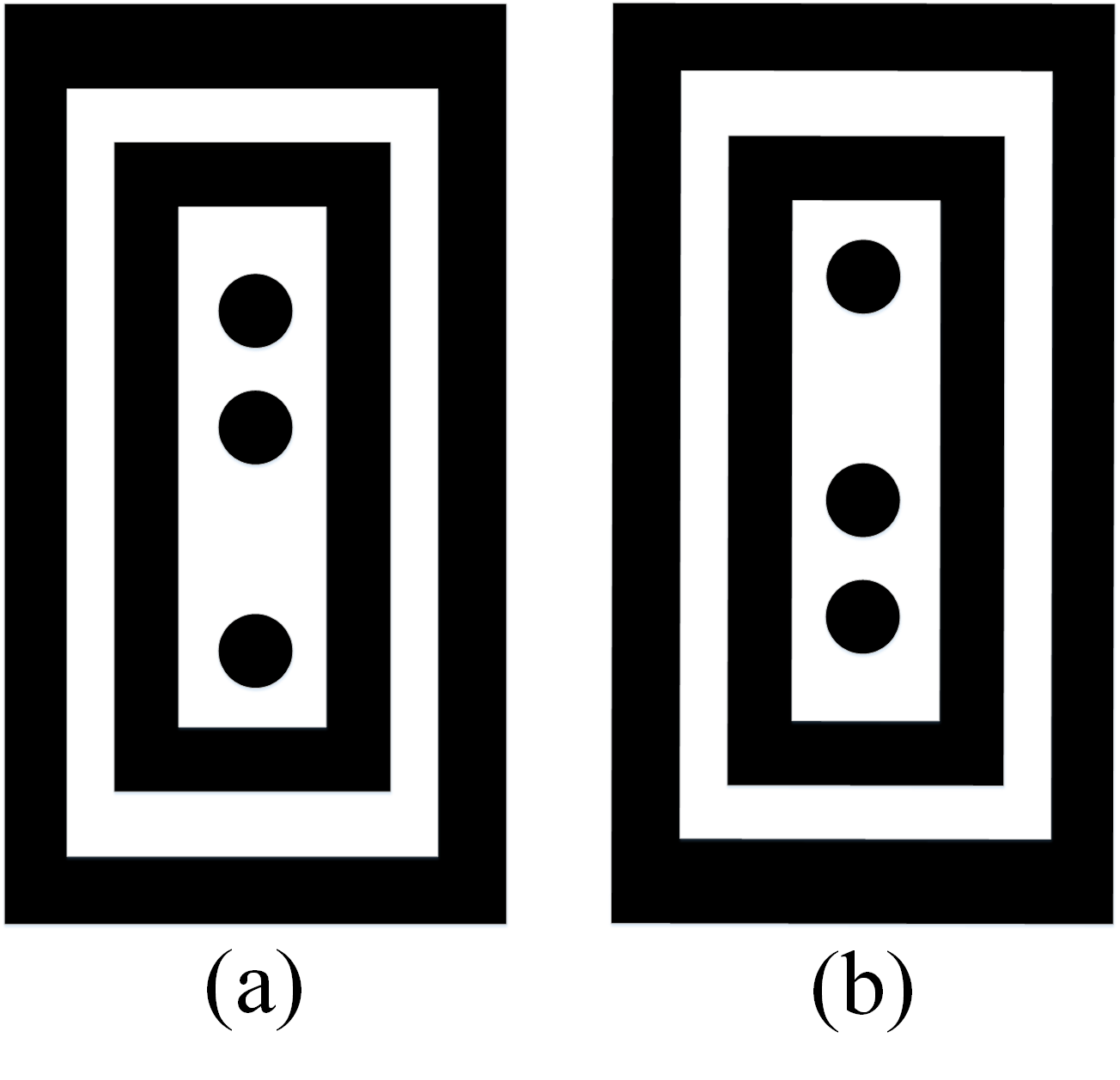}
\caption{`Slalom' pattern. (a) Turn right for certain degrees; (b) Turn left for certain degrees.}
\label{fig_slalompattern}
\end{figure}

To further explore the agility of the robot system based on the SCAMP-5, we also proposed a `slalom' application with some pre-designed visual patterns (see Fig. \ref{fig_slalompattern}) encoding turning and angle information which are extracted using the SCAMP-5 vision system. By placing these patterns in specific positions, the rover weaves left and right around the patterns according to the information encoded on each, in a similar manner to slalom skiing. The image processing method is similar to the one used for four disks extraction.

\begin{figure}[t!]
\centering
\includegraphics[width=3.3in]{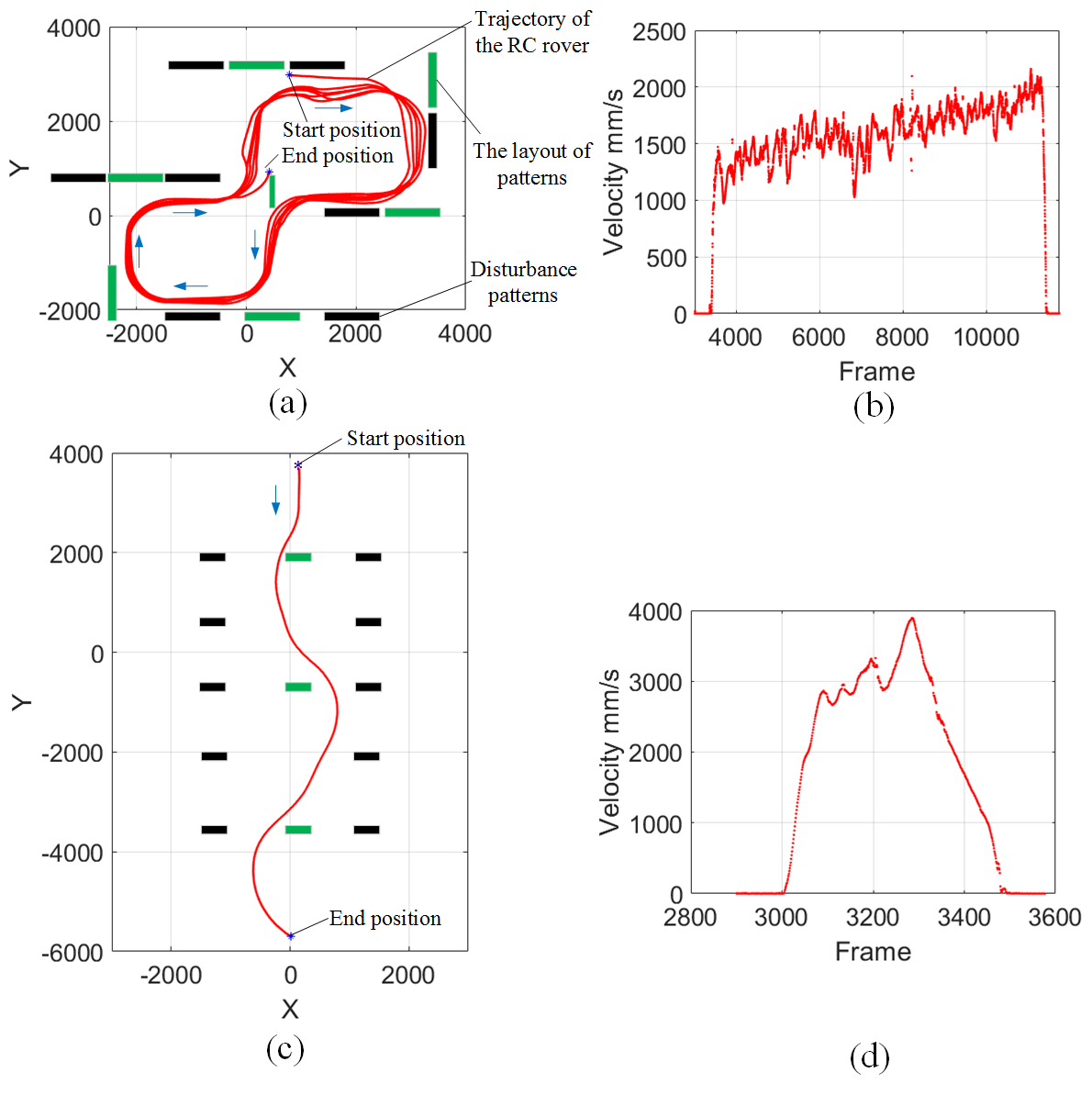}
\caption{Trajectory and velocity curve for the `slalom' process. (a) trajectory of number `8'; (b) velocity curve for number `8'; (c) trajectory of a fast `slalom'; (d) velocity curve for the fast `slalom'.}
\label{fig_slalom}
\end{figure}

Fig. \ref{fig_slalom} shows the trajectory and velocity of the rover during navigation. In Fig. \ref{fig_slalom}(a), there are eight patterns placed in the arena, six of them guide the rover to turn right and two to turn left. The robot turns then continues towards the next closest pattern it encounters. Fig. \ref{fig_slalom}(b) shows the average speed of 1.70 m/s for this scenario. By placing these gates in a line, a high-speed slalom can be achieved. In Fig. \ref{fig_slalom}(c), the distance between gates are set 2.40 m apart, and the rover turns right or left at a distance 0.80 m from the pattern. Its velocity curve (see Fig. \ref{fig_slalom}(d)) shows the highest speed reaches 3.88 m/s which translates to 139 km/h for the real-sized car. 

\section{Conclusion and Future Work}

This paper focuses on real-time visual information processing using a Pixel Processor Array and agile robot navigation with information derived from the visual sensor. The SCAMP-5 is capable of processing high-frame rate visual information as the agile navigation requires [10]. In this paper, we implemented the high-speed reactive navigation and enable a Subaru mobile robot to run through multiple gates at an average speed of 2.20 m/s and a highest speed of 3.88 m/s for a `slalom'. Moreover, the developed vision algorithms based on the SCAMP-5 
can reach up to 2000 fps with enough illumination. We proposed a direct link between the image information and the angle of the steerable servo motor in the mobile robot. To the best of our knowledge, the speed of the agile ground vehicle achieved is the highest among the systems using non-conventional visual-processor pipelines.

The multi-gate test shows high-speed navigation for an agile reactive navigation system and shows the effectiveness of the control method under motion disturbance and vibrations. However, the navigation speed and performance is still limited by many factors including low control frequency of the servo motor or loss of targets especially when the fast-moving mobile robot is far from the target. In our future work, we aim to improve the system adaptability in both indoor and outdoor environments. 
We believe that the development of 
novel vision processing hardware architectures
is key to the progress of agile and responsive robotics that are required to operate in a complex and uncertain world. This work explores the case of agile reactive navigation and our future emphasis will include extending the capabilities beyond known target detection and into classification and place recognition while navigating.

\section{Data Access Statement and Acknowledgements}\label{sec11}

This work was supported by UK EPSRC EP/M019454/1, EP/M019284/1, EPSRC Centre for Doctoral Training in Future Autonomous and Robotic Systems: FARSCOPE and China Scholarship Council (No. 201700260083). Experiments are performed at the Bristol Robotics Laboratory. The nature of the task and PPA means that the SCAMP-5 images in this work are not recorded.

\bibliographystyle{ieeetr}
\bibliography{IET-Submission}

\end{document}